\documentclass[runningheads]{llncs}

\usepackage{eccv}

\usepackage{eccvabbrv}

\usepackage{graphicx}
\usepackage{booktabs}
\usepackage{wrapfig}
\usepackage[accsupp]{axessibility}  

\usepackage{hyperref}

\usepackage{orcidlink}
\usepackage{algorithm}
\usepackage{algorithmic}
\usepackage{multirow}
\usepackage{nicematrix}

\begin{document}

\title{Neural Spectral Decomposition for Dataset Distillation}


\author{Shaolei Yang\inst{1} \and
Shen Cheng\inst{2} \and
Mingbo Hong\inst{2} \and
Haoqiang Fan\inst{2} \and
Xing Wei\inst{1} \and
Shuaicheng Liu\inst{2,3,\dagger}}

\authorrunning{S. Yang et al.}

\institute{School of Software Engineering, Xi’an Jiaotong University, Xi’an, China \\
\email{\{yangshaolei@stu.xjtu.edu.cn, weixing@mail.xjtu.edu.cn\}} \\
\and Megvii Technology, Beijing, China \\
\email{\{chengshen,mingbohong97,fhq\}@megvii.com} \\
\and University of Electronic Science and Technology of China, Chengdu, China \\
\email{liushuaicheng@uestc.edu.cn}\\
$^\dagger$Corresponding Author
}

\maketitle

\begin{center}
  \centering
  \captionsetup{type=figure}
  \includegraphics[width=1\textwidth]{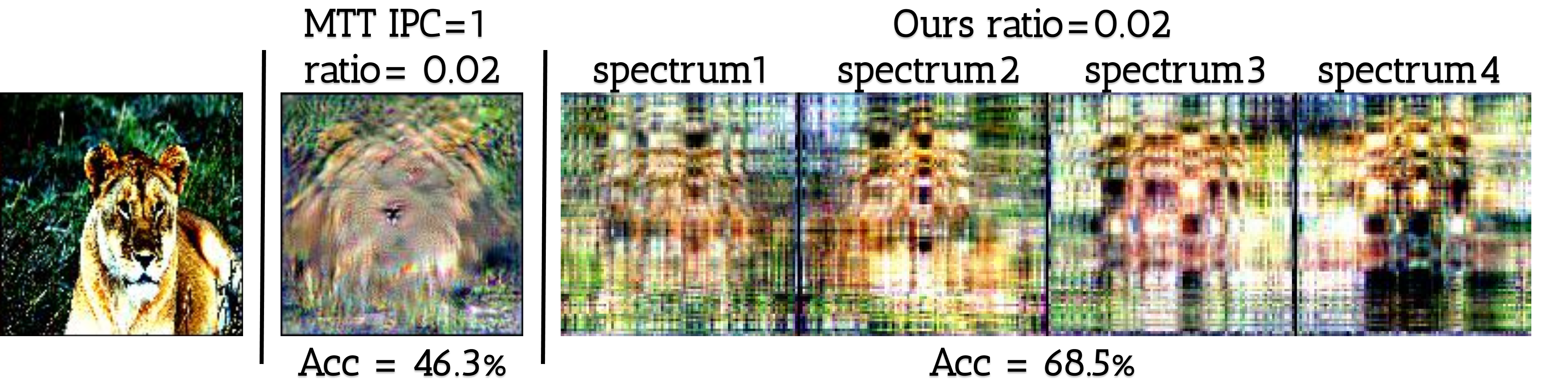}
  \caption{\textit{Left}: An example image belonging to the class `lion'. \textit{Middle}: MTT achieves an accuracy of 46.3\% with a compression ratio of 0.02. \textit{Right}: The proposed method enhances the accuracy to 68.5\% under the same compression ratio.} 
  \label{fig:teaser}
\end{center}%

\begin{abstract}

In this paper, we propose Neural Spectrum Decomposition, a generic decomposition framework for dataset distillation. Unlike previous methods, we consider the entire dataset as a high-dimensional observation that is low-rank across all dimensions. We aim to discover the low-rank representation of the entire dataset and perform distillation efficiently. Toward this end, we learn a set of spectrum tensors and transformation matrices, which, through simple matrix multiplication, reconstruct the data distribution. Specifically, a spectrum tensor can be mapped back to the image space by a transformation matrix, and efficient information sharing during the distillation learning process is achieved through pairwise combinations of different spectrum vectors and transformation matrices. Furthermore, we integrate a trajectory matching optimization method guided by a real distribution. Our experimental results demonstrate that our approach achieves state-of-the-art performance on benchmarks, including CIFAR10, CIFAR100, Tiny Imagenet, and ImageNet Subset. Our code are available at \url{https://github.com/slyang2021/NSD}.
\end{abstract}
\section{Introduction}
Deep learning models have achieved remarkable success across a wide range of real-world applications ~\cite{sam,track,gpt,caption, dino}, primarily owing to the availability of extensive datasets. These large-scale datasets empower models with enhanced generalization abilities, scalable expansion potential, and heightened accuracy. However, training models on such massive datasets presents significant challenges. Primarily, the vast volume of data presents considerable challenges in terms of storage, transmission, and data preprocessing. Moreover, publishing raw data inevitably gives rise to practical concerns regarding privacy and copyright issues. To address these issues, dataset distillation~\cite{random1, Herding1, forget,DM,DC,IDC,MTT,ma2024curriculumdatasetdistillation, li2024single} techniques have emerged, aiming to distill the knowledge contained within large-scale datasets into smaller, synthetic ones. This approach reduces the storage while ensuring that the model captures consistent information from both the original and distillate datasets. One classical approach is coreset selection~\cite{random1, Herding1, forget}, which aims to identify a smaller subset from the original dataset that can preserve the essential features and structure while offering approximate or accurate results. Another highly efficient method involves the generation of synthetic data. For instance, DM (Distribution Matching) ~\cite{DM} minimizes the distribution discrepancy between real and synthetic data. DC (Dataset Condense) ~\cite{DC} matches the training gradients, while MTT (matching training trajectory) ~\cite{MTT} emphasizes aligning the parameter trajectories throughout the training process. These strategies, namely optimization-based methods, enable the compression of large-scale datasets while maintaining the essential information required for effective learning and inference in deep learning models.


However, these methods distillate directly in the image space, lacking the fact that natural images satisfy regularity conditions that form a low-rank data subspace. IDC ~\cite{IDC} has been introduced to generate more informative images and effectively utilize the condensed data elements. Specifically, IDC divides images into patches and treats each patch as an individual image, enabling learning within the low-frequency subspace of the image space. This approach allows for leveraging the inherent structure present in natural images. More recently, an advancement~\cite{memory} has been made to calculate shared memory across data. The concept of shared memory can be understood as subspaces spanning across the dimensions of the dataset. A step further, generator-based methods ~\cite{haba,mttgan,WeiCYM23,gdz} leverage additional prior knowledge to enhance the realism of synthetic images and effectively constrain the optimization direction. By incorporating the prior knowledge of the generator, the generated images closely resemble real-world examples. We note the above methods as parameterization techniques.

Unfortunately, these parameterization methods still have some limitations, such as the requirement for additional storage space, data, and even training costs. Our key observation is that the success of parameterization methods relies on 1) low-rankness and 2) information sharing. For example, Deng et al ~\cite{memory} demonstrates that the information between different samples is low-rank, allowing for more samples to be enhanced within the same storage space, while the generator-based approach also shares the parameter space among different samples, making distillation more effective.

In this work, we propose Neural Spectrum Decomposition, which innovatively decomposes the entire dataset into high-dimensional spectrum tensors and kernel transformation matrices. The information expressed in each dimension of this tensor is low-rank, allowing for further reduction in storage space. Furthermore, we create a series of spectrum vectors and transformation matrices that can generate a large number of samples through pairwise combinations. This decomposition enables information sharing across dimensions, making distillation learning more efficient and enhancing the informativeness of the samples. As illustrated in Fig.~\ref{fig:teaser}, the network trained with this decomposition approach has a more discriminative feature space.

In terms of optimization methods, we employ a trajectory matching strategy guided by real samples, where the trajectory matching is augmented with guidance from the real dataset distribution to compensate for the limitations of trajectory sampling. In our experiments, the results demonstrate a 47.9\% improvement over the baseline method based on trajectory matching on CIFAR10 under the condition of $IPC=1$. In addition, we achieve state-of-the-art performance on benchmarks, including CIFAR10, CIFAR100, and TinyImageNet. In summary, our contributions can be summarized as follows:

\begin{itemize}
    \item  We propose a generic decomposition framework based on two principles: low-rankness and information sharing.
    \item We introduce an optimization method that improves trajectory matching.
    \item Our experiments demonstrate the state-of-the-art (SOTA) performance of our method, and we provide an in-depth analysis of our approach.
\end{itemize}

\section{Related Work}
    \label{sec:related}
	\noindent\textbf{Dataset Condensation.} 
            \cite{dd} proposed for the first time dataset distillation (DD), which aims to synthesize small datasets so that the models trained using the small datasets have similar performance to the original dataset. Since then, a lot of interesting work has emerged. \cite{DC} proposed a dataset condensation(DC) method based on gradient matching, which formulates dataset condensation as a gradient matching problem between the gradients of the deep neural network weights trained on the original and synthetic data. Subsequently, \cite{DSA} proposed Differentiable Siamese Augmentation(DSA), which can be effectively used to synthesize more informative synthetic images using data augmentation, resulting in better performance when using augmentation to train the network. Because previous methods focused on individual steps and were unable to model complete trajectories, \cite{MTT} proposed MTT to match parametric trajectory segments trained on synthetic data with expert trajectory segments from models trained on real data, thus avoiding short-sightedness. Trajectory matching methods suffer from the so-called cumulative trajectory error caused by the discrepancy between distillation and subsequent evaluation. To mitigate the adverse effects of this cumulative trajectory error, \cite{FTD} proposed FTD to encourage optimization methods to seek flat trajectories.
    
            Despite the good performance of gradient-matching-based methods, the computational cost of their synthesis is still high due to the complex two-layer optimization and second-order derivative computation. \cite{DM} proposed DM to synthesize condensed images by matching the feature distributions of the synthesized and original training images. Thus it can be applied to more realistic and larger datasets with complex neural architectures. \cite{CAFE} proposed CAFE, which can characterize the distribution of the original samples well by aligning layer-by-layer features between real and synthetic data while explicitly encoding discriminative capabilities into the synthetic set, and introducing a new two-layer optimization.

	\noindent\textbf{Efficient Distilled Dataset Parameterization.}
            Unlike the above-mentioned methods for generating representative images, other methods focus on Dataset Parametrization. \cite{IDC} proposed IDC for efficient parameterization of synthetic data, where four synthetic images are downsampled and thus stitched together into a single image under the same storage budget and recovered using bilinear interpolation in downstream training. \cite{haba} proposed HaBa, a further optimization of IDC, to divide the dataset into two components: data Hallucination networks and Bases, where the latter is fed into the former to reconstruct image samples.  \cite{memory} achieve more efficient and effective distillation through shared common representation. Concretely, they learn a set of bases (aka "memories") which are shared between classes and combined through learned flexible addressing functions to generate a diverse set of training examples. \cite{mttgan} proposed GLaD, which employs a deep generative prior by parameterizing the synthetic dataset in the intermediate feature space of generative models. Although these methods achieve better data compression rates as well as higher accuracy, they are not conducive to migration to downstream tasks because they usually need to store additional information to generate images.
            
\noindent\textbf{Spectral Decomposition.} Spectral decomposition can convert images from the spatial domain to the spectrum, and process the image by analyzing the information in the spectrum. \cite{dec1} found that noise levels in different spectral components have different contrasts, and the target was recovered from the low-frequency components of the image and then enhanced image details from the high-frequency components.  
\cite{dec2} proposes a learning-based frequency selection method to remove redundant frequency components without losing accuracy.\cite{dec3} designed an adaptive frequency band filtering operator to filter the characteristics of different frequency bands through position-by-position multiplication operations, achieving a better trade-off between accuracy and efficiency. In this work, we decompose the image to achieve cross-dimensional information sharing, so that the distilled image can better learn the features of different frequencies.
\begin{figure*}[t]
    \centering
    \includegraphics[width=1.\linewidth]{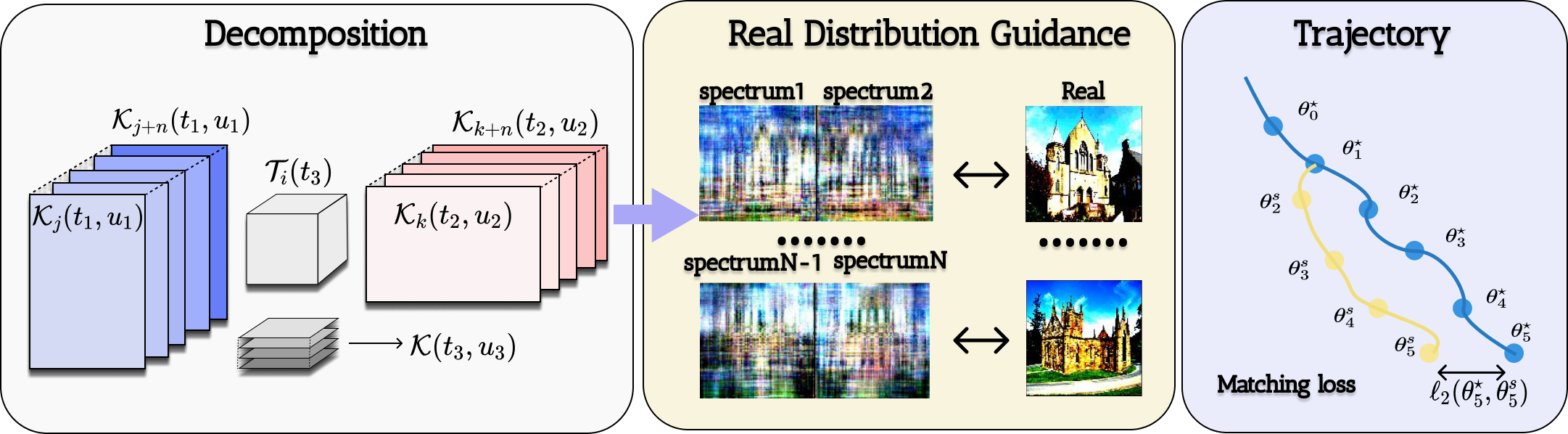}
    
    \caption{The overall pipeline. The core lies in the decomposition approach of the image generation component. In terms of training strategy, in addition to using trajectory matching, we also incorporate guidance from the distribution of real samples.}
    \label{fig:enter-label}
\end{figure*}
\section{Method}

In this section, we describe the proposed method where each image can be decomposed into a spectrum tensor and a transformation matrix. The entire synthetic dataset $\mathcal{D}_{syn}$ is a series of spectrum tensors and transformation matrices combinations.  From this viewpoint, we can employ a flexible image decomposition approach, leading to improved performance through the utilization of a diminished parameter quantity. Fig.~\ref{fig:enter-label} illustrates our overall pipeline.
In the following, we first give a general definition of the decomposition
and then we step by step analyze the details of our method.

\paragraph{\textbf{Formulation}.} Given a target dataset $\mathcal{D}_{syn} = \{ \boldsymbol{x}_{i}\}^{N}_{i=1}$, we aim to find a decomposition in which each image $x_{i}$ can be represented by a spectrum tensor $\mathcal{T}(t_{n}) \in \mathbb{R}^{t_{1} \times t_{2} \ldots \times t_{n}} $ and a kernel matrix $\mathcal{K}(t_{n}, u_{n}) \in \mathbb{R}^{t_{1} \times t_{2} \ldots \times t_{n} \times u_{1} \times u_{2} \ldots \times u_{n}}$, where $n$ is the number of dimensions, $t_{n}$ and $u_{n}$ represent the original and transformed length of dimension $n$, respectively. By performing a simple matrix multiplication, the kernel matrix can map the length $t_n$ of each dimension of $\mathcal{T}$ to $u_n$.Then we define a generic decomposition as:
\begin{equation}
    \begin{aligned}
        \boldsymbol{x}_{j+(i-1)N_{\mathcal{K}}} = \mathcal{T}_{i}(t_{n}) \mathcal{K}_j(t_{n}, u_{n}),   \\
        \textrm{s.t.}  \quad 1 \leq i \leq N_{\mathcal{T}},  1 \leq j \leq N_{\mathcal{K}},   \label{con:imgsyn}
    \end{aligned}
\end{equation}
where $N_{\mathcal{T}}, N_{\mathcal{K}}$ denote the numbers of spectrum tensors and transformation matrices respectively. Here, $j+(i-1)N_{\mathcal{K}}$ is the index of each image which is computed by the $i$-th $\mathcal{T}_{i}$ and the $j$-th $\mathcal{K}_{j}$. 

For each dimension, $t_n$ is projected to $u_n$. The compression behavior is due to $t_n \le u_n$. As $\frac{t_{n}}{u_{n}}$ decreases, the compression rate increases; conversely, as $\frac{t_{n}}{u_{n}}$ increases, the compression rate decreases. Inspired by popular deep-learning frameworks, we simply set $\mathcal{T}$ as a 4-dimensional tensor (similar to BCHW format), namely $\mathcal{T} \in \mathbb{R}^{t_{1} \times t_{2} \times t_{3} \times t_{4}}$. However, even for a 4-dimensional tensor, the space required to store the kernel matrix is quite significant. To this end, we treat the transformations of each dimension as independent. After this decomposition, we can use any existing optimization framework to learn $\mathcal{T}$ and $\mathcal{K}$. This decomposition can be viewed as an implicit structural constraint.

\paragraph{\textbf{Separable Kernel}.} As different dimensions are independent of each other, the entire kernel matrix can be decomposed into multiple sub-kernel matrices, which are as follows:
\begin{equation}
    \mathcal{K}(t_n, u_n) =  \mathcal{\hat{K}}(t_1, u_1) \ldots \mathcal{\hat{K}}(t_n, u_n),
\end{equation}
where $\mathcal{\hat{K}}(t_n, u_n) \in \mathbb{R}^{t_{n} \times u_{n}}$. By performing dimension decomposition, the number of parameters can be further reduced, decreasing from $\prod t_n u_n$ to $\sum t_n u_n$.

\paragraph{\textbf{Combination}.} For each spectrum tensor, it can be combined with different kernel matrices. Similarly, the same kernel matrix can also be combined with different spectrum tensors. This combination strategy increases the exposure of $\mathcal{T}$ and $\mathcal{K}$ in each sampling process, thereby improving training efficiency. This information sharing across different dimensions can be seen as a form of low-rankness. By using this combination, we obtain more diverse samples.
\begin{algorithm}[tb]
   \caption{}
   \label{alg:alg}
   \begin{algorithmic}[1]
   \STATE {\bfseries input:} Real Dataset $\mathcal{D}_{real}$;
   \STATE {\bfseries input:} Spectrum Tensor $\mathcal{T}$, Kernel Matrix $\mathcal{K}$; 
   \STATE {\bfseries input:} Distillation Step N, Expert Trajectory Length M; 
   \STATE {\bfseries input:} Expert Weights  $\left\{\mathcal{\theta}^{t}\right\}$,  learning rate $\alpha$;
   \STATE {\bfseries input:} loss function $norm-2$ $\mathcal{\ell(\cdot,\cdot)}$ and cross-entropy $L$
   \REPEAT
   \STATE Obtain synthetic dataset $\mathcal{D}_{syn}$ with eqn.(\ref{con:imgsyn}) \\
   \STATE Initialize student weight: $\mathcal{\theta}^{s}_{i} := \mathcal{\theta}^{t}_{i}$ \\
   \FOR{$t=1$ {\bfseries to} $N$}
   \STATE Sample a mini-batch $B \subset \mathcal{D}_{syn}$
   \STATE Compute gradients $g_{L}=\nabla_\theta L_{B}({\theta}^{s}_{i+t-1})$
   \STATE Update student weight: ${\theta}^{s}_{i+t}={\theta}^{s}_{i+t-1} - \alpha{g_{L}}$
   \ENDFOR
   \STATE Compute match loss $ \ell({\theta}^{s}_{i+N},{\theta}^{t}_{i+M})$ \\
   \STATE \textbf{Compute real guided loss $ L({\theta}^{s}_{i+N}, \mathcal{D}_{real})$} \\
   \STATE Update  $\mathcal{T}$ and $\mathcal{K}$ with respect to $\gamma L$ + $\ell$ \\
   \UNTIL{Converge}
\end{algorithmic}
\end{algorithm}

\paragraph{\textbf{Training Strategy}.} This decomposition can be plugged into any optimization algorithm. Building upon the advantage of long-term information, we enhance the performance by using MTT~\cite{MTT} (matching training trajectory) as a baseline and subsequently incorporating distribution information into the optimization. A trajectory contains the parameters which are trained with $T$ steps on real or synthetic data, denoted as $\tau^*=\left\{\theta_t^*\right\}_0^T$ or $\tau^s=\left\{\theta_t^s\right\}_0^T$.  The objective is to minimize $\left\|\theta^{s}_{t+N}-\theta_{t+M}^*\right\|_2^2$. However, the effect of distillation depends on how many teacher trajectories are available, and it is difficult for teacher trajectories to cover the full parameter space of possibilities. Hence, extracting information from real data distributions to guide optimization can enhance performance. One intuition to use real data information is that \textsl{networks trained on synthetic data can classify correctly in real data}, which is similar to KIP ~\cite{KIP}. Assuming the final parameter of the synthesized trajectory is ${\theta}^{s}_{i+N}$, we sample a batch of data from real trajectories and compute a classification loss for this data $ L({\theta}^{s}_{i+N}, \mathcal{D}_{real})$, as depicted in Alg.~\ref{alg:alg}.

\begin{table*}[h]\footnotesize
	 \caption{\small Top-1 test accuracy of test models comparison to state-of-the-art methods. We use ConvNet3 for training and testing and show the mean and standard deviation of the five tests. We compared the coreset selection method, the distilled dataset parametrization method, and the original method(optimized in the original image space) separately. IPC: Images Per Class(budget follows Eqn. \ref{con:budget}), Ratio (\%): the ratio
of condensed images to the whole training set.} 
        
	\begin{center}
	    \small
        \resizebox{1.\linewidth}{!}{
		\begin{NiceTabular}{c|c|ccc|cc|c}
			\toprule
	\multirow{1}{*}{} & 		  
     Dataset &  \multicolumn{3}{c}{CIFAR10} & \multicolumn{2}{c}{CIFAR100} & \multicolumn{1}{c}{TinyImageNet}  \\
                \midrule
                \multirow{2}{*}{} &
			   IPC & 1 & 10 & 50 & 1 & 10 & 1  \\
			   & Ratio\% & 0.02 & 0.2 & 1 & 0.2 & 2 & 0.2   \\
			\hline
                \multirow{3}{*}{Coreset} &
			Random
			& 14.4\tiny$\pm$2.0
			& 26.0\tiny$\pm$1.2
			& 43.4\tiny$\pm$1.0
			& 4.2\tiny$\pm$0.3
			& 14.6\tiny$\pm$0.5
			& 1.4\tiny$\pm$0.1
			\\
			& Herding
			& 21.5\tiny$\pm$1.2
			& 31.6\tiny$\pm$0.7
			& 40.4\tiny$\pm$0.6
			& 8.4\tiny$\pm$0.3
			& 17.3\tiny$\pm$0.3
			& 2.8\tiny$\pm$0.2
			\\
			& Forgetting
			& 13.5\tiny$\pm$1.2
			& 23.3\tiny$\pm$1.0
			& 23.3\tiny$\pm$1.1
			& 4.5\tiny$\pm$0.2
			& 15.1\tiny$\pm$0.3
			& 1.6\tiny$\pm$0.1
			\\
                \midrule
   \multirow{3}{*}{Parametrization} &
			IDC~\cite{IDC}
			& 50.0\tiny$\pm$0.4
			& 67.5\tiny$\pm$0.5
			& 74.5\tiny$\pm$0.1
			& 25.1\tiny$\pm$0.3
			& 44.8\tiny$\pm$0.2
			& -
			\\
			
                &HaBa~\cite{haba}
			& 48.3\tiny$\pm$0.8
			& 69.9\tiny$\pm$0.4
			& 74.0\tiny$\pm$0.2
			& 33.4\tiny$\pm$0.4
			& 40.2\tiny$\pm$0.2
			& -
			\\
                &Deng et al.~\cite{memory}
			& 66.4\tiny$\pm$0.4
			& 71.2\tiny$\pm$0.4
			& 73.6\tiny$\pm$0.5
			& 34.0\tiny$\pm$0.4
			& 42.9\tiny$\pm$0.7
			& 16.0\tiny$\pm$0.7
			\\
                \midrule
                \multirow{7}{*}{Original} &
			DC~\cite{DC}
			& 28.3\tiny$\pm$0.5
			& 44.9\tiny$\pm$0.5
			& 53.9\tiny$\pm$0.5
			& 12.8\tiny$\pm$0.3
			& 25.2\tiny$\pm$0.3
			& -
			\\
			&DSA~\cite{DSA}
			& 28.8\tiny$\pm$0.7
			& 52.1\tiny$\pm$0.5
			& 60.6\tiny$\pm$0.5
			& 13.9\tiny$\pm$0.3
			& 32.3\tiny$\pm$0.3
			& -
			\\
			&DM~\cite{DM}
			& 26.0\tiny$\pm$0.8
			& 48.9\tiny$\pm$0.6
			& 63.0\tiny$\pm$0.4
			& 11.4\tiny$\pm$0.3
			& 29.7\tiny$\pm$0.3
			& 3.9\tiny$\pm$0.2
			\\
			&KIP to NN~\cite{KIP}
			& 49.9\tiny$\pm$0.2
			& 62.7\tiny$\pm$0.3
			& 68.6\tiny$\pm$0.2
			& 15.7\tiny$\pm$0.2 
			& 28.3\tiny$\pm$0.1
			& -
			\\
			&CAFE + DSA~\cite{CAFE}
			& 31.6\tiny$\pm$0.8
			& 50.9\tiny$\pm$0.5
			& 62.3\tiny$\pm$0.4
			& 14.0\tiny$\pm$0.3
			& 31.5\tiny$\pm$0.2
			& -
			\\
			
                &FRePo~\cite{FRePo}
			& 46.8\tiny$\pm$0.7
			& 65.5\tiny$\pm$0.4
			& 71.7\tiny$\pm$0.2
			& 28.7\tiny$\pm$0.1
			& 42.5\tiny$\pm$0.2
			& 15.4\tiny$\pm$0.3
			\\
                &FTD~\cite{FTD}
			& 46.8\tiny$\pm$0.3
			& 66.6\tiny$\pm$0.3
			& 73.8\tiny$\pm$0.2
			& 25.2\tiny$\pm$0.2
			& 43.4\tiny$\pm$0.3
			& 10.4\tiny$\pm$0.3
			\\
                &MTT~\cite{MTT}
			& 46.3\tiny$\pm$0.8
			& 65.3\tiny$\pm$0.7
			& 71.6\tiny$\pm$0.2
			& 24.3\tiny$\pm$0.3
			& 40.1\tiny$\pm$0.4
			& 8.8\tiny$\pm$0.3
			\\
                \midrule
                \multirow{1}{*}{} 
   			&\textbf{ours}
			& \textbf{68.5\tiny$\pm$0.8}
			& \textbf{73.4\tiny$\pm$0.2}
			& \textbf{75.2\tiny$\pm$0.6}
			& \textbf{36.5\tiny$\pm$0.3}
			& \textbf{46.1\tiny$\pm$0.2}
			& \textbf{21.3\tiny$\pm$0.2}
                \\
                \midrule
                \multirow{1}{*}{} &
			Full dataset
			& \multicolumn{3}{c}{84.8\tiny$\pm$0.1} 
			& \multicolumn{2}{c}{56.2\tiny$\pm$0.3}
    		& \multicolumn{1}{c}{37.6\tiny$\pm$0.4} 
			\\
			\bottomrule
		\end{NiceTabular}
		}
	\end{center}
	\small

 \label{tbl:main_acc}
\end{table*}
\section{Experiments}
    \label{sec:experiment}
    In this section, we first introduce the datasets used and the specific experimental details. Next, we compare the proposed method with the state-of-the-art methods. Subsequently, we also conduct the ablation experiments and give an in-depth analysis. Finally, we will visualize our condensed dataset and show the impact of the decomposition.

    
    \subsection{Datasets \& Implementation Details}
    \noindent\textbf{CIFAR-10/100} The \textit{CIFAR-10/100}~\cite{CIFAR} consists of 60,000 32x32 color nature images, each with 50,000 training images and 10,000 test images equally divided into 10/100 categories.
    
    \noindent\textbf{TinyImageNet} The \textit{TinyImageNet}~\cite{TinyImageNet} dataset contains 200 categories, each with 500 training images, 50 validation images, and 50 test images of 64×64 pixels, and the images are from the ImageNet dataset~\cite{imagenet}.
    
    \noindent\textbf{ImageNet Subset} The ImageNet~\cite{imagenet} dataset is a large-scale and diverse image database containing over 14 million images. It is utilized for training and evaluating computer vision algorithms. Next, we run our method on a subset of the ImageNet, which is divided into 6 subsets following the MTT~\cite{MTT}, namely ImageNette (various objects), ImageWoof (dog breeds), ImageFruit (fruits), ImageMeow (cats), ImageSquawk (birds) and ImageYellow (yellow objects).Each subset contains more than 10,000 images, and we adjusted the resolution of all images to 128×128 according to the original settings.

    \noindent\textbf{Implementation details.} 
    For a fair comparison, we adopt the same parameter settings as previous work, and in all datasets, the amount of condensed image parameters is the same as when Images Per Class (IPC) is 1/10/50, respectively. The model uses a simple three-layer convolutional neural network, each layer consists of a 3×3 convolution kernel, a convolutional layer with 128 channels, Instance normalization, RELU, and 2×2 average pooling with stride 2. Following MTT~\cite{MTT}, 5 experiments are performed during evaluation, and the mean and variance of 5 experiments are finally taken. The learning rate is 0.01, epoch is 1000, batchsize is 256, momentum is 0.9, weight decay is 0.0005, and the default data enhancement is the same as DSA~\cite{DSA}. Our experiments were run on eight RTX2080ti GPUs.

    
    
    \noindent\textbf{Storage budgets.} 
    Because our method saves the spectrum tensor and kernel matrix rather than the synthesized image directly, we ensure that our compression rate is roughly equal to what it would be if IPC = 1/10/50, i.e.
        
    \begin{equation}
        \begin{aligned}
            size(Spectrum ~Tensor) + size(Kernel ~Matrix) \\ 
            \approx IPC \times C \times size(Image)
             \label{con:budget}
        \end{aligned}
    \end{equation}
    where $size(\cdot)$ denotes the size of the occupied storage and $C$ denotes the number of classes. 

    \noindent\textbf{Model details.} 
    We compare our method with the baseline method in the eqn. \ref{con:budget} premise. Specifically, we make $t_3=t_4$ and $t_2=3$ in all experiments. When $IPC=1$ and $IPC=10$, we take $t_3,t_4$ to be one-half of the original image resolution, and when $IPC>10$, we take $t_3, t_4$ to be seven-eighths of the original image size. Finally, the choice of $t_1$ and $u_1$ depends on the storage budget, and in principle we make $t_1 < u_1$. All the synthetic images are trained for 10k iterations and the spectrum tensor and kernel matrix are used with the SGD optimizer, We set the momentum rate to 0.9, and the spectrum tensor and kernel matrix are initialized with random initialization.

\begin{table*}[ht!]\tiny
    \tiny
	 \caption{\small We show the test results of different decomposition methods on the distillation dataset, and all of the decomposition methods achieve higher accuracy gains compared to baseline, and the learnable 
 kernel matrix method also performs better relative to the fixed kernel matrix.} 
        
	\begin{center}
	    
        \resizebox{1.\linewidth}{!}{
		\begin{tabular}{c|cc|ccccccc}
			\toprule
     Dataset & IPC & Ratio\% & BaseLine & DWT & SVD & LSVD & DCT &LDCT & ours 
                 \\ \midrule
                \multirow{2}{*}{CIFAR10} &
                 1 & 0.02 & 46.3\tiny$\pm$0.8 & 51.3\tiny$\pm$0.6 & 49.0\tiny$\pm$0.6 
                 & 53.8\tiny$\pm$0.2 & 57.9\tiny$\pm$0.3 & 62.3\tiny$\pm$0.4 & 68.5\tiny$\pm$0.8 \\
                 & 10 & 0.2 & 65.3\tiny$\pm$0.7 & 66.9\tiny$\pm$0.7 & 67.9\tiny$\pm$0.3
                 & 69.4\tiny$\pm$0.5 & 69.5\tiny$\pm$0.2 & 69.9\tiny$\pm$0.5 & 73.4\tiny$\pm$0.2
                 \\ \midrule
                \multirow{1}{*}{CIFAR100} &
                1 & 0.2 & 24.3\tiny$\pm$0.3 & 26.4\tiny$\pm$0.4 & 25.3\tiny$\pm$0.4 
                & 27.7\tiny$\pm$0.3 & 30.1\tiny$\pm$0.3 & 35.5\tiny$\pm$0.2 & 36.5\tiny$\pm$0.3 \\
                
			\bottomrule
		\end{tabular}
		}
	\end{center}
	\tiny

 \label{tbl:transfer}
\end{table*}
    \subsection{Evaluation on the Condensed Data}
        In this subsection, we evaluated the performance of our method on low-resolution classification datasets and the higher-resolution ImageNet dataset. Compared to previous methods, our approach achieved state-of-the-art results.
        
        \noindent\textbf{Low-Resolution Data:} We compare our method to three state-of-the-art dataset distillation methods on various benchmark datasets~\cite{CIFAR, TinyImageNet}.
        First, we compare our method with the three core subset selection methods, Random~\cite{random1}, Herding~\cite{Herding1}, and Forgetting~\cite{forget}. Second, we compare our method with the original method(optimized in the original image space) for comparison, including Dataset Condensation(DC)~\cite{DC},  Differentiable Siamese Augmentation (DSA)~\cite{DSA}, Distribution Matching(DM)~\cite{DM}, Kernel-Inducing-Point(KIP)~\cite{KIP}, Aligning Features(CAFE)~\cite{CAFE}, Flat Trajectory Distillation(FTD)~\cite{FTD}, Feature Regression with Poolin(FRePo)~\cite{FRePo}, Matching Training Trajectories(MTT)~\cite{MTT}. Finally, we compare our method with the Distilled Dataset Parametrization method, including Information-intensive Dataset Condensation(IDC)~\cite{IDC}, Hallucination networks and Bases(HaBa)~\cite{haba}, Addressable Memories~\cite{memory}. We present the performance of our method and the method mentioned above in Table~\ref{tbl:main_acc}, our method achieves state-of-the-art on all benchmark datasets compared to the previous three types of methods. In particular, We can readily observe that our method obtains a great performance gain relative to the original method, especially when the synthesized data is fewer. In CIFAR10 with IPC=1, we obtain a 37.4\% improvement in accuracy relative to the previous state-of-the-art method, and even on  CIFAR100 and TinyImageNet, we obtain 26.1\% and 22.7\% improvements, respectively. In addition, our method also obtains consistency improvement relative to the parametrization method, by 3.2\%, 6.4\%, and 18.1\% on the CIFAR10, CIFAR100, and TinyImageNet datasets, respectively, with IPC = 1. 


\begin{table}[ht!]
\tiny
  \caption{Performance comparison of our method with other methods on ImageNet Subsets with IPC=1.}
\centering
\small
\setlength{\tabcolsep}{1.5mm}{
  \begin{tabular}{c|cccc|c}
   \hline
   Dataset & MTT~\cite{MTT} & HaBa~\cite{haba} & GLaD~\cite{mttgan} & ours & Full Dataset \\
   \hline
ImageFruit&26.6$\pm$0.8&34.7$\pm$1.1&23.1$\pm$0.4&\textbf{39.8$\pm$0.2}&63.9$\pm$2.0\\
ImageNette&47.7$\pm$0.9&51.9$\pm$1.6&38.7$\pm$1.6&\textbf{68.6$\pm$0.8}&87.4$\pm$1.0\\
ImageWoof&28.6$\pm$0.8&32.4$\pm$0.7&23.4$\pm$1.1&\textbf{35.2$\pm$0.4}&67.0$\pm$1.3\\
ImageSquawk&37.3$\pm$0.8&41.9$\pm$1.4&35.8$\pm$1.4&\textbf{52.9$\pm$0.7}&87.5$\pm$0.3\\
ImageMeow&26.6$\pm$0.4&36.9$\pm$0.9&26.0$\pm$1.1&\textbf{45.2$\pm$0.1}&66.7$\pm$1.1\\
ImageYellow&45.2$\pm$0.8&50.4$\pm$1.6&26.6$\pm$0.4&\textbf{61.0$\pm$0.5}&84.4$\pm$0.6\\
\hline
     \end{tabular}}


  \label{tab:rebuttal_imagenet}
\end{table}

        \noindent\textbf{ImageNet Subsets:} We also experimented our approach on the more challenging ImageNet~\cite{imagenet} dataset, following the MTT~\cite{MTT} setup, we additionally extended the distillation model from the original three-layer ConvNet to a five-layer ConvNet and evaluated it on six ImageNet Subsets, and the test results are displayed in Table~\ref{tab:rebuttal_imagenet}. An astute observation reveals our method distinctly outperforms all others in every experiment, simultaneously achieving a remarkable improvement over the baseline method. Most notably, on the ImageNette dataset, we attained a staggering 43\% enhancement in accuracy compared to MTT\cite{MTT}. This unequivocally substantiates the efficacy of our method and further corroborates its adeptness at more profoundly understanding and enhancing the informative value of samples.

    \subsection{Cross-architecture generalization}

        To facilitate more effective application to other task scenarios, the synthetic dataset must demonstrate the ability to generalize across different architectures. However many of the current dataset distillation methods suffer from performance degradation when synthetic datasets are trained on one network structure and tested on another.
        
        To evaluate the ability of our method to generalize on other structures, we distill the CIFAR10 on ConvNet(3-layer)~\cite{convnet} with IPC=10, then we use four different structures to do the evaluation, ConvNet3~\cite{convnet}, ResNet18~\cite{resnet}, VGG11~\cite{vgg}, AlexNet~\cite{alexnet}.Table~\ref{tbl:cross} shows the results of the generalization performance of our method on different structures, and our method achieves better generalization performance compared to previous methods(DC\cite{DC},KIP\cite{KIP},MTT\cite{MTT},FRePo\cite{FRePo},Deng et al.\cite{memory}). In particular, our method exceeds Deng et al. \cite{memory} by 3.7\%, 1.6\%, 11.8\%, 11.9\% on ConvNet3, ResNet18, VGG11, AlexNet, respectively. In addition, our method only drops 7.5, 4.6, and 1.0 on ResNet18, VGG11, and AlexNet, respectively. Especially on AlexNet, our method has almost no performance degradation. Because our method learns the decomposed spectrum tensor and kernel matrix, our method can learn the information of each frequency band well, and the cross-structural generalizability experiments also prove that our method has excellent generalizability.
 

\begin{table}[ht!]
	
\caption{Cross-Architecture Results trained with ConvNet on  CIFAR-10 with $\texttt{ipc}=  10$. We report test results of our method and previous state-of-the-art methods on ConvNet, ResNet18, VGG11, AlexNet.}
\centering
\small
\setlength{\tabcolsep}{3.5mm}{
  \begin{tabular}{c|cccc}
   \toprule
&\multicolumn{4}{c}{Evaluation Model}\\
   Method&ConvNet&ResNet18&VGG11&AlexNet\\
   \midrule
DC~\cite{DC}&44.9$\pm$0.5&18.4$\pm$0.4&35.9$\pm$0.7&22.4$\pm$1.4\\
KIP~\cite{KIP}&47.6$\pm$0.9&36.8$\pm$1.0&42.1$\pm$0.4&24.4$\pm$3.9\\
MTT~\cite{MTT}&65.3$\pm$0.7&46.6$\pm$0.6&50.3$\pm$0.8&34.2$\pm$2.6\\
FRePo~\cite{FRePo}&65.5$\pm$0.4&58.1$\pm$0.6&59.4$\pm$0.7&61.9$\pm$0.7\\
Deng et al.\cite{memory}&71.2$\pm$0.4& 63.9$\pm$0.3 & 60.6$\pm$0.2 &63.8$\pm$0.8\\
\textbf{ours}&\textbf{73.4$\pm$0.2}&\textbf{64.9$\pm$1.3}&\textbf{67.8$\pm$0.2}&\textbf{71.4$\pm$0.3}\\
\bottomrule
     \end{tabular}}


    \label{tbl:cross}
\end{table}

    \subsection{Exploring Spectral Decomposition}
        We also explored different variants of neural spectral decomposition. Regarding the selection of different $\mathcal{K}$, we have discrete wavelet transform (DWT), singular value decomposition (SVD), and discrete cosine transform (DCT). 
        

        \noindent\textbf{Discrete Wavelet Transform.}
        The discrete wavelet transform is specifically designed to emphasize the high-frequency components of an image, capturing intricate details and making it highly suitable for image compression. On the other hand, the wavelet transform considers both frequency and spatial information, effectively preserving the spatial properties of the image. We use the haar format as our base for the wavelet transform to categorize the image into LL,LH,HL,HH where L denotes low frequency and H denotes high-frequency. We randomly sample some frequency bands at a time during the inverse transform, so that we can better learn the information of each band during the distillation.

        \noindent\textbf{Singular Value Decomposition.}
        Since singular value decomposition is a generalization of the theory of spectral analysis to arbitrary matrices, singular value decomposition can be well applied to signal processing, statistics and other fields. The formula for singular value decomposition is $M=U\Sigma V$, where the columns of U and V are the left and right singular vectors in the singular values, respectively, and the elements on the diagonal of the matrix $\Sigma$ are equal to the singular values of M. Specifically, we transform the picture from $(B, C, H, W)$ to $(B, C \times H \times W)$, and then perform singular value decomposition on the transformed matrix. After transformation, the spectrum tensor is $V$, and the kernel matrix is $U\Sigma$(in order to better compress the image, we only take the first $n$ rows of $V$ as well as the first $n$ columns of $U$, and the number of parameters follows eqn. \ref{con:budget}), then we only optimize the spectrum tensor, when synthesizing the dataset.

        For the sake of general discussion, we introduce learnable SVD(LSVD), which optimizes the kernel matrix along with the spectrum tensor optimization.
        
        \noindent\textbf{Discrete Cosine Transform.}
        The discrete cosine transform is often used by signal processing and image processing because the discrete cosine transform can bring together the more important information of an image in one piece.  It is commonly utilized for lossy data compression of signals and various types of images, including still and moving images.  Specifically, we first construct two transformation matrices  $\mathcal{K}_l, \mathcal{K}_r$ using the discrete cosine transform, where
         $\mathcal{K}_l \in \mathbb{R}^{m \times n}, \mathcal{K}_r \in \mathbb{R}^{n \times m}$.Then we construct two inverse transform matrices $\mathcal{IK}_l, \mathcal{IK}_r$,  where
         $\mathcal{IK}_l \in \mathbb{R}^{n \times m}, \mathcal{IK}_r \in \mathbb{R}^{m \times n}$. Therefore spectrum tensor $\mathcal{T} = \mathcal{K}_l \times \mathcal{O}_{img} \times \mathcal{K}_r$, where  $\mathcal{O}_{img} \in \mathbb{R}^{3 \times n \times n}$ denotes the original image. The number of parameters follows eqn. \ref{con:budget}. The discrete cosine transform formula is shown below:
        \begin{equation}
        \begin{aligned}
            \mathcal{K}_{l_{i,j}} = cos(\frac{\pi}{n} \times (j+0.5) \times (i+0.5)), \\
            \textrm{s.t.}  \quad 1 \leq i \leq m,  1 \leq j \leq n, 
        \label{con:DCT}
        \end{aligned}
        \end{equation}
        $\mathcal{K}_r$ is calculated in the same way as in eqn. \ref{con:DCT}, and the inverse transformation only needs to be multiplied by $\frac{2}{n}, \frac{2}{m}$ for pairs of $\mathcal{K}_l$ and $\mathcal{K}_r$, respectively.

        In the meantime, we introduce the learnable discrete cosine transform (LDCT), i.e., we also learn the kernel matrix, which allows the synthesized image to better utilize the information on different time domains. 

        Table~\ref{tbl:transfer} shows the test results between the datasets distilled by different decomposition methods, the baseline is MTT~\cite{MTT}.It's easy to see that the original image after Spectral Decomposition can be better utilized for the information in the time domain, so that the synthesized dataset not only can obtain the low-frequency information but also can be better obtained to the high-frequency information. Meanwhile, the learnable kernel matrix has higher accuracy than the fixed kernel matrix, which indicates that the learnable kernel matrix is also more efficient in utilizing the information in different frequency bands.
        
    \begin{figure}[!t]
\centering

\includegraphics[width=0.8\linewidth]{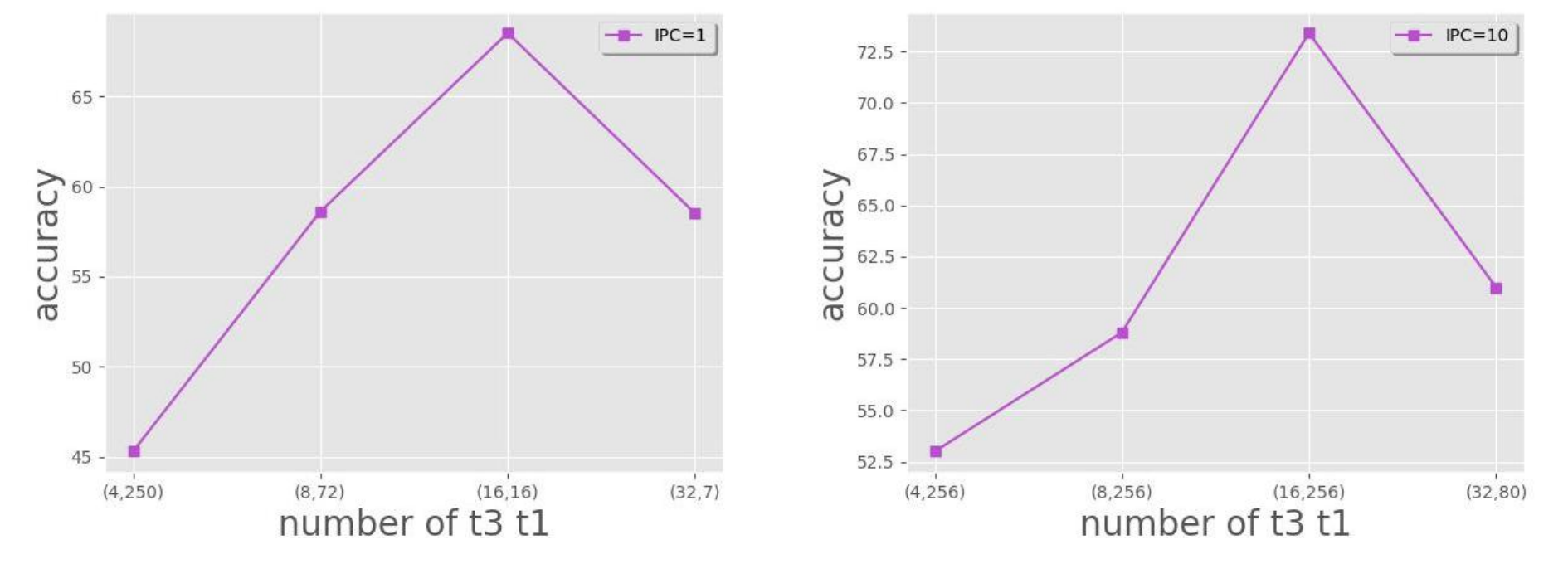} 
\caption{Accuracy at different number(t3,t1) on the CIFAR10 dataset with IPC=1/10.}

\label{abla_t}
\end{figure}
    \begin{table}[ht!]
  \caption{
Test accuracy of our method on different baseline methods with (CIFAR10,IPC=10).}
\centering
\small
\setlength{\tabcolsep}{2.5mm}{
  \begin{tabular}{c|cccc}
   \toprule
&\multicolumn{4}{c}{Evaluation Model}\\
   Method&ConvNet&ResNet18&VGG11&AlexNet\\
   \midrule
DC~\cite{DC}&44.9$\pm$0.5&18.4$\pm$0.4&35.9$\pm$0.7&22.4$\pm$1.4\\
w. ours&55.6$\pm$0.3&29.5$\pm$1.1&40.1$\pm$0.3&33.5$\pm$0.2\\
Gain &10.7&11.1&4.2&11.1\\
\midrule
DM~\cite{DM}&48.9$\pm$0.6&36.8$\pm$1.2&41.2$\pm$1.8&34.9$\pm$1.1\\
w. ours&50.1$\pm$0.3&38.9$\pm$0.4&42.5$\pm$0.5&46.2$\pm$0.3\\
Gain &1.2&2.1&1.3&11.3\\
\midrule
MTT~\cite{MTT}&65.3$\pm$0.7&46.6$\pm$0.6&50.3$\pm$0.8&34.2$\pm$2.6\\
w. ours&73.4$\pm$0.2&64.9$\pm$1.3&67.8$\pm$0.2&71.4$\pm$0.3\\
Gain &8.5&18.3&17.5&37.2\\
\bottomrule
     \end{tabular}}

  \label{tab:diff_base}
\end{table}

    \subsection{Ablation Studies}
    In this subsection, we study ablations to investigate the effectiveness of each module and the influence of the hyper parameters.
    
    \noindent\textbf{Building upon Different Baselines.}
    To demonstrate the pervasiveness of our method, we implement our method on the training pipeline of multiple state-of-the-art methods, including DC~\cite{DC}, DM~\cite{DM}, and MTT~\cite{MTT}, and we evaluate the performance of the synthetic dataset on the CIFAR10 dataset and with IPC=10. As Table~\ref{tab:diff_base} demonstrates, our method achieves significant improvements on the baseline method when both training and testing on ConvNet. Specifically, our method improves 23.8\%, 10.6\%, and 13.0\% on ConvNet with respect to DC, DM, and MTT, respectively, and for cross-structure testing, our method improves even more for accuracy than on ConvNet. This shows that our method not only migrates well to other methods but also leads to better generalization of synthetic datasets.

    \noindent\textbf{Number of $t$ .}
    We further investigate the effects of $t_1$, $t_2$ and $t_3$, and Figure \ref{abla_t} presents the test accuracy of the synthetic dataset with different $t_1$ and $t_3$ choices. We default $t_3 = t_4$, keeping the same size scale as the original image. We can discover that it is more beneficial to learn the synthetic dataset when t3 takes a moderate size. 
    


\begin{table*}
\caption{Ablation study on different modules of our method with CIFAR10 dataset. Dec: Spectral Decomposition, Guided loss: Real guided loss, Acc: Top-1 test accuracy .}
\centering
\small
\centering
\setlength{\tabcolsep}{3.5mm}
\begin{tabular}{c|c|c|c|c}
        \toprule
         MTT~\cite{MTT} & Dec  &  Guided loss  &IPC & Acc\\
         \midrule
         \checkmark & &  & 1 & 46.3$\pm$0.8\\
         \checkmark & \checkmark &  & 1 & 67.9$\pm$0.5(\textbf{$\uparrow$ 21.6})\\
         \checkmark & \checkmark & \checkmark & 1 & 68.5$\pm$0.8(\textbf{$\uparrow$ 22.2})\\
         
         \hline
         \checkmark & &  & 10 & 65.3$\pm$0.7\\
         \checkmark & \checkmark &  & 10 & 71.8$\pm$0.3(\textbf{$\uparrow$ 6.5})\\
         \checkmark & \checkmark & \checkmark & 10 & 73.4$\pm$0.2(\textbf{$\uparrow$ 8.1})\\
         
         \bottomrule
        \end{tabular}

\label{tab:gloss}
\end{table*}

    \noindent\textbf{Real Distribution Guidance.}
    We conduct an ablation study in Table~\ref{tab:gloss} to verify the effectiveness of our method, we add our module on baseline(MTT~\cite{MTT}) respectively, since MTT learns the long-term information of the expert model well, the addition of the Decomposition module is a great improvement for the accuracy, which is improved by 46.6\%( cifar10,ipc=1), at the same time, because MTT does not have constraints on the student model on real data and, so the addition of Real guided loss also improved by 0.8\%. This result validates the effectiveness of our method.

    \subsection{Visualizations}

    In this subsection, we visualize the inter-dimensional similarity to show the effectiveness of our method.
    

    \begin{figure}[ht]
\centering
\includegraphics[width=1\linewidth]{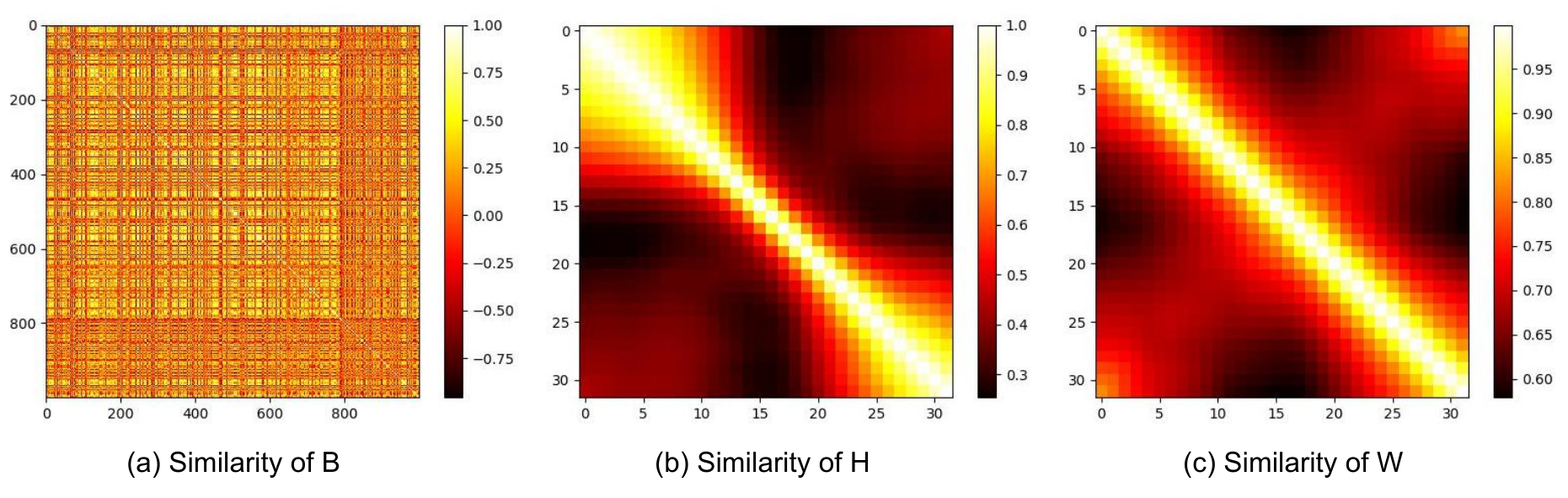} 
\caption{We compute the similarity between B-dimension, H-dimension, and W-dimension, respectively, for the original graph with dimensions BCHW.}

\label{similary_BCNW}
\end{figure}

     \noindent\textbf{Similarity.}
     In Figure~\ref{similary_BCNW}, a subset comprising 1,000 images is randomly sampled from the airplane category within the CIFAR10~\cite{CIFAR} dataset. Subsequently, we conduct an analysis to compute the inter-dimensional similarity across the Batch (B), Height (H), and Width (W) dimensions within the context of the original image tensor format, denoted as BCHW. It is imperative to note that the analysis reveals a discernible local correlation among the dimensions. This observed correlation suggests the presence of redundant information along these axes, which is indicative of the underlying low-rank structure of the image data. The low-rank characteristic of an image implies that the image can be represented by a smaller number of features than that of the original dimensionality without significant loss of information. This is due to the inherent redundancy present within the image; for instance, neighboring pixels often share similar color values and patterns, leading to this correlation.
\section{Conclusion}
    \label{sec:conclusion}

In this study, we primarily revisit parameterization methods in dataset distillation and summarize two key points for the success of these methods: low-rank and information sharing. In contrast to previous methods that only leverage specific dimensional information, such as low-frequency information from individual images, the proposed method fully utilizes the low-rankness across the entire dataset. Additionally, the combination of different spectral tensors and kernel matrices effectively enhances learning efficiency. Considering the effectiveness of low-rank, we believe that there is significant potential for exploring the proposed method in high-resolution image learning in the future.

\noindent{\bf Acknowledgement}: 
This work was supported by National Key R\&D Program of China (No. 2021ZD0110400), the Fundamental Research Funds for the Central Universities (No. xxj032023020), National Natural Science Foundation of China (No.62372091) and Sichuan Science and Technology Program of China (No.2023NSFSC0462), and sponsored by the CAAI-MindSpore Open Fund, developed on OpenI Community.


%
%
\bibliographystyle{splncs04}
\bibliography{main}
\end{document}


\title{Neural Spectral Decomposition for Dataset Distillation}


\author{Shaolei Yang\inst{1} \and
Shen Cheng\inst{2} \and
Mingbo Hong\inst{2} \and
Haoqiang Fan\inst{2} \and
Xing Wei\inst{1} \and
Shuaicheng Liu\inst{2,3,\dagger}}

\authorrunning{S. Yang et al.}

\institute{School of Software Engineering, Xi’an Jiaotong University, Xi’an, China \\
\email{\{yangshaolei@stu.xjtu.edu.cn, weixing@mail.xjtu.edu.cn\}} \\
\and Megvii Technology, Beijing, China \\
\email{\{chengshen,mingbohong97,fhq\}@megvii.com} \\
\and University of Electronic Science and Technology of China, Chengdu, China \\
\email{liushuaicheng@uestc.edu.cn}\\
$^\dagger$Corresponding Author
}

\maketitle

In the Supplementary Material, we will further analyze and, secondly, we will show some other visualization results.


\begin{figure}[ht]
\centering
\includegraphics[width=0.5\linewidth]{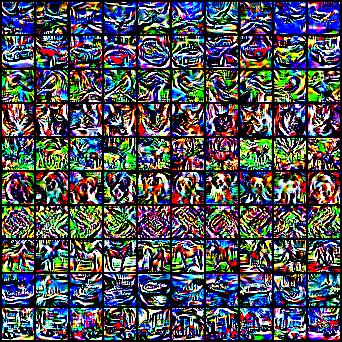} 
\caption{Visualization results of synthetic images with (CFIAR10, Ratio=0.2) on MTT.}

\label{mtt}
\end{figure}
\begin{figure*}[!t]
\centering
\includegraphics[width=1\linewidth]{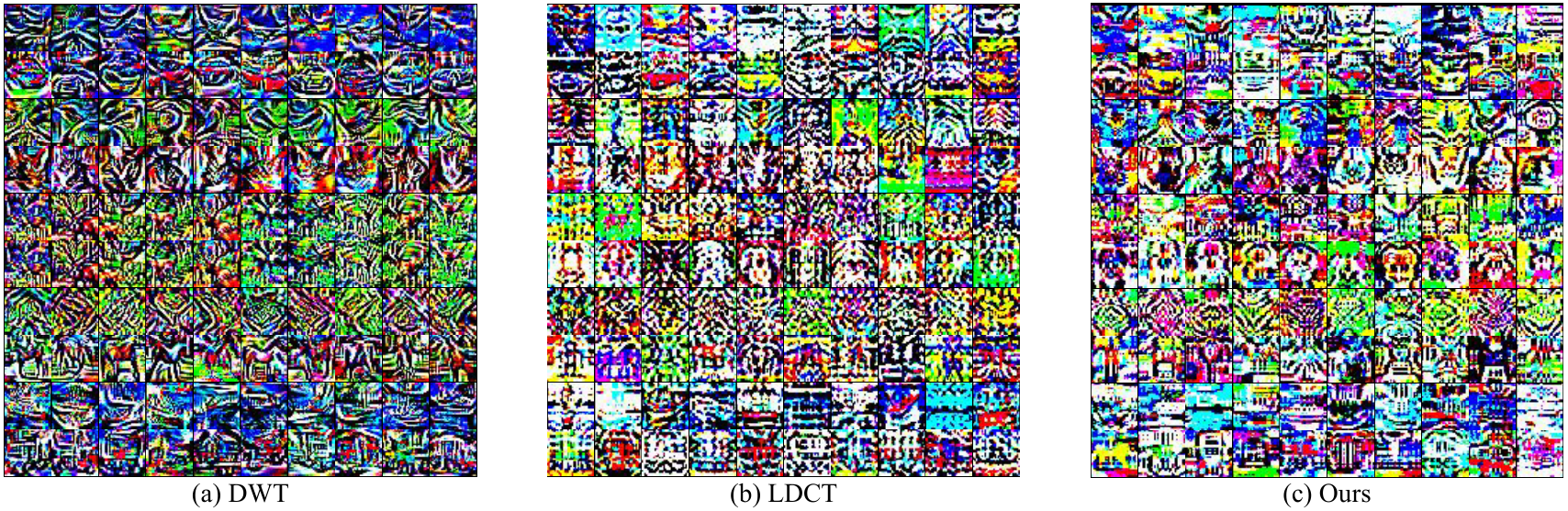} 
\caption{Visualization results of synthetic images generated by using different kernel form. (a) DWT, (b) LDCT and (c) ours on CIFAR10 with IPC=10.}

\label{syn_img}
\end{figure*}

We further visualize other synthetic images, Figure \ref{mtt} shows the visualization results of MTT on CIFAR10, our method can make better use of the information between the spectra and can make use of the high-frequency information compared to MTT. 

\begin{figure*}[hbp]
\centering
\includegraphics[width=0.6\linewidth]{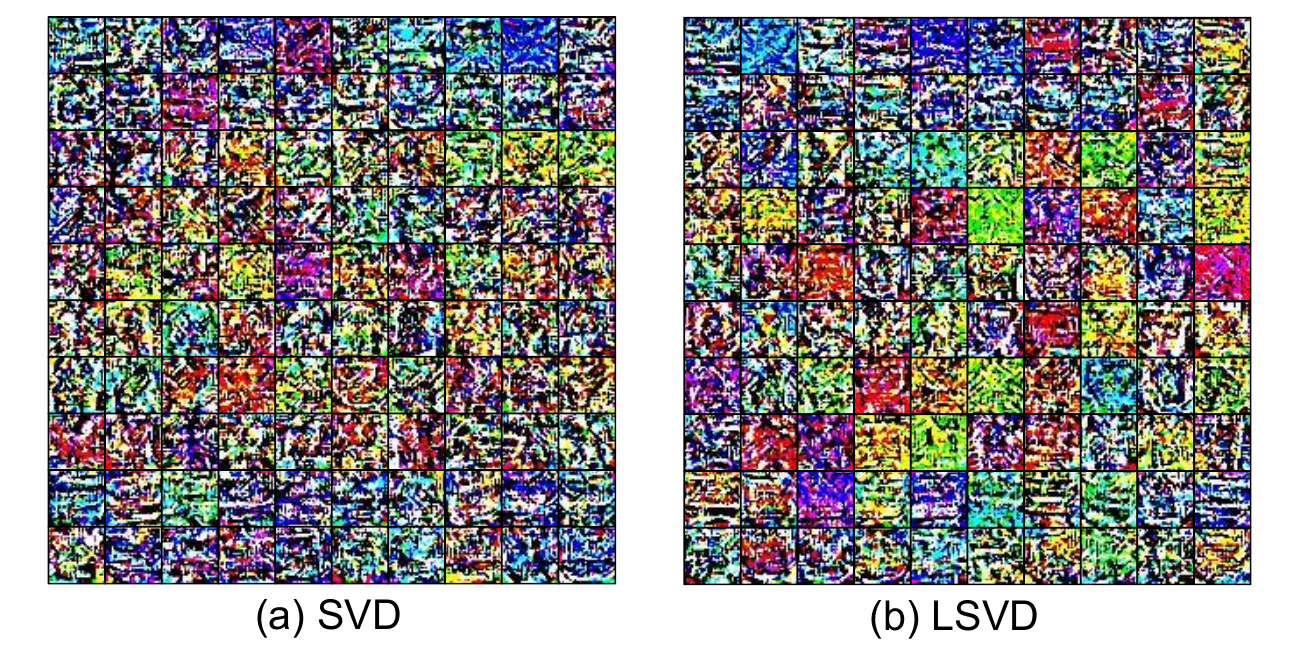} 
\caption{Visualization results of synthetic images generated by using different kernel form. (a) SVD, (b) LSVD on (CIFAR10, Ratio=0.02)}

\label{svd}
\end{figure*}

\begin{figure*}[ht]
\centering
\includegraphics[width=0.9\linewidth]{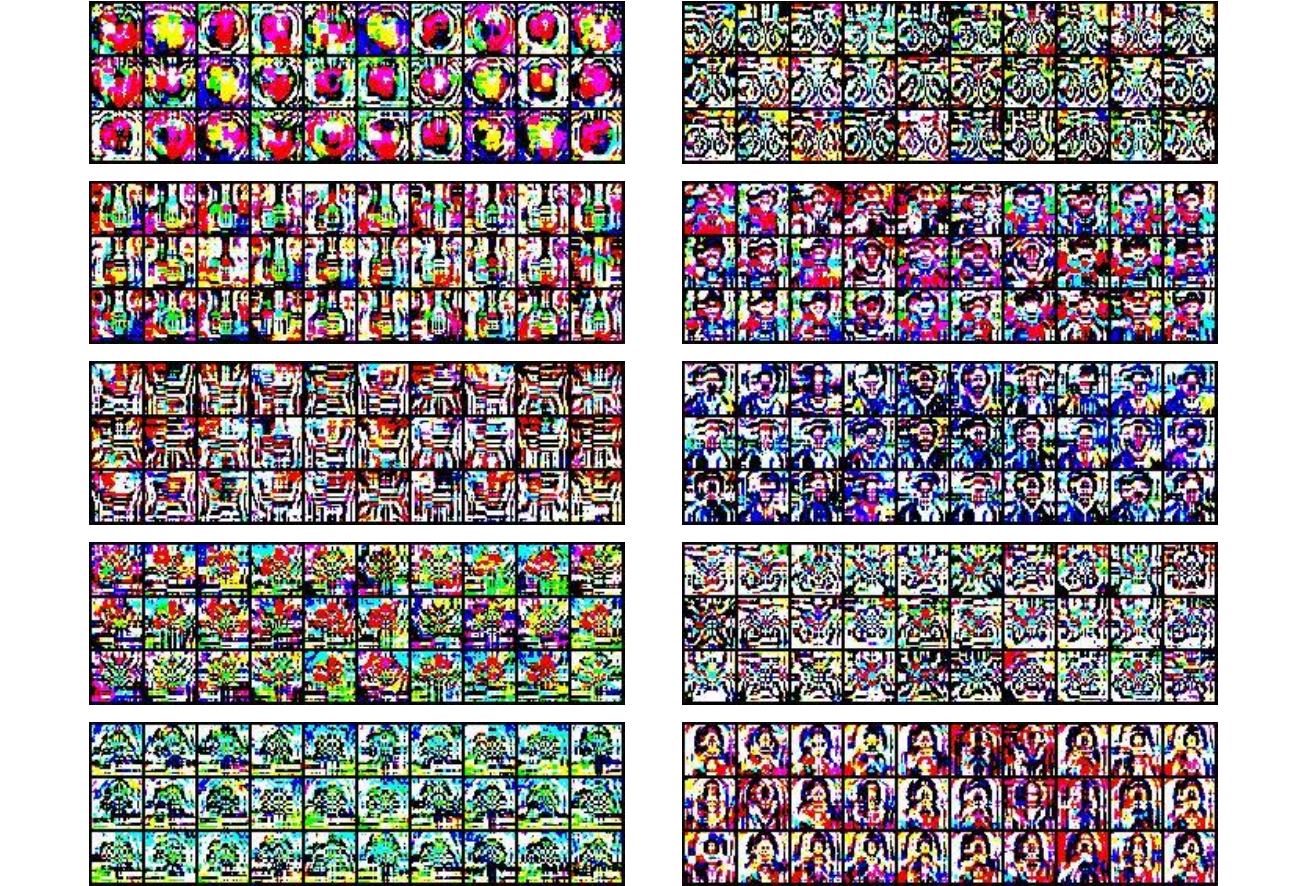} 
\caption{Visualization results of synthetic images with (CFIAR100, Ratio=0.2).}

\label{cifar100}
\end{figure*}

\begin{figure*}[ht]
\centering
\includegraphics[width=0.75\linewidth]{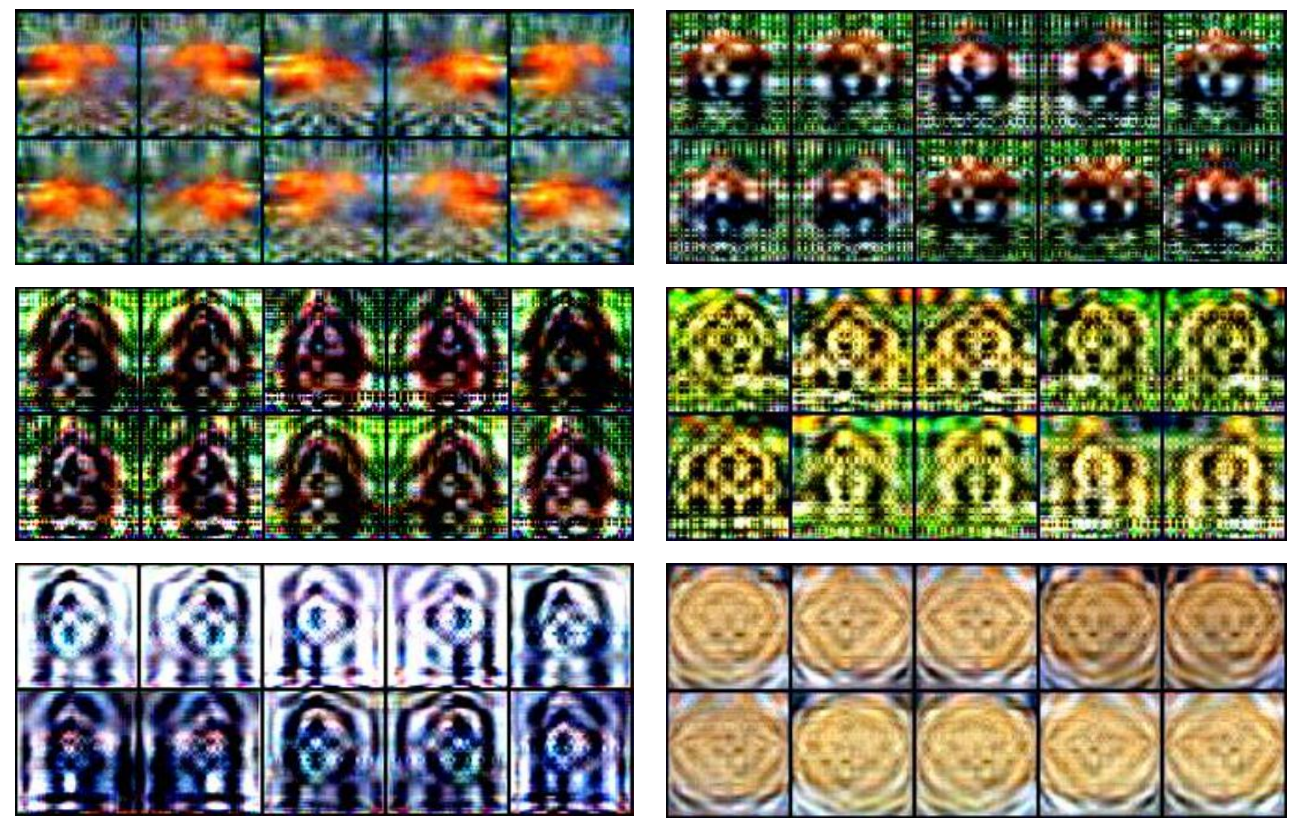} 
\caption{Visualization results of synthetic images with (TinyImageNet, Ratio=0.2).}

\label{tiny}
\end{figure*}

\begin{figure*}[ht]
\centering
\vspace{-0.1in}
\includegraphics[width=0.9\linewidth]{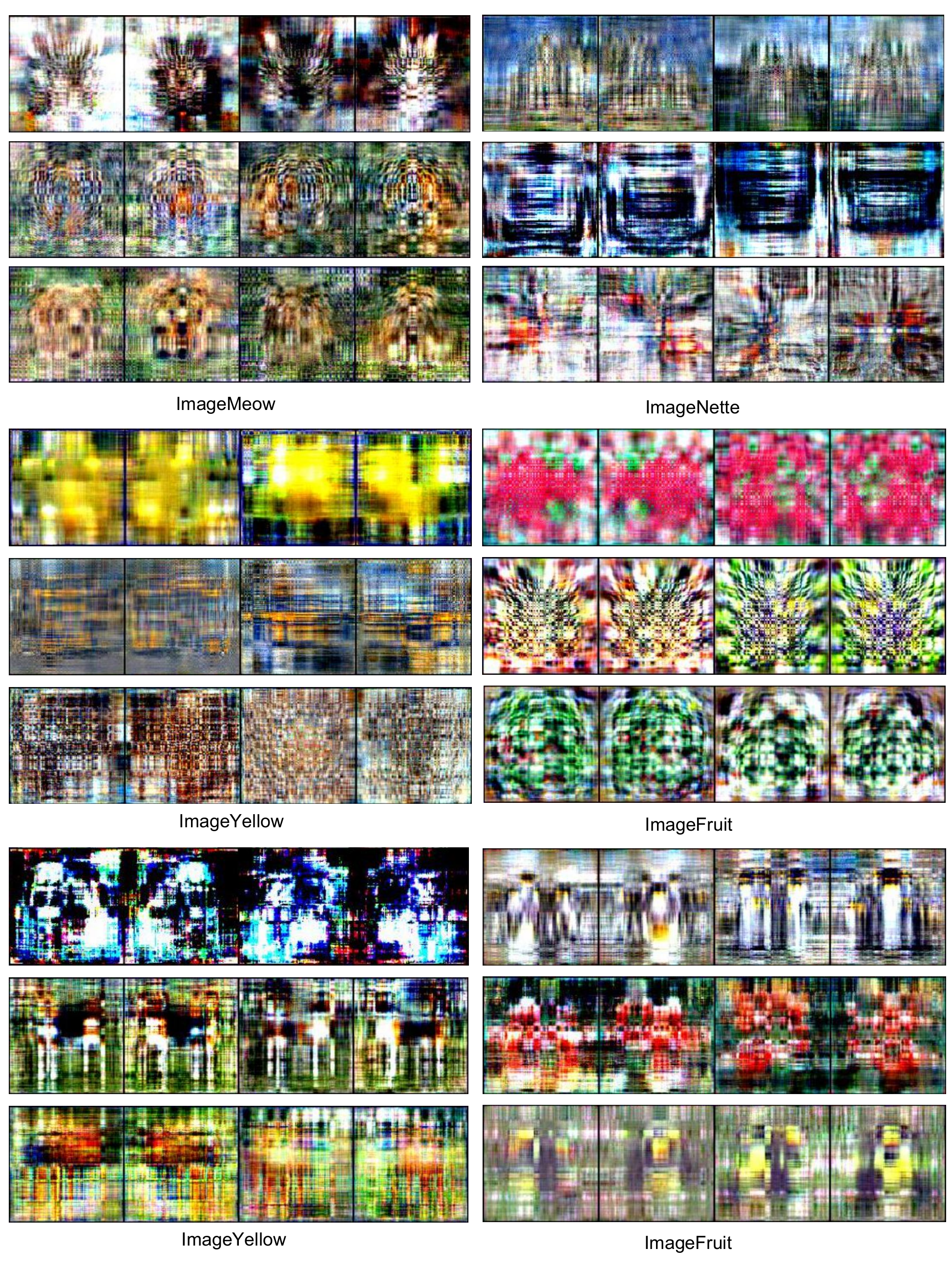} 
\caption{Visualization results of synthetic images with (ImageNet Subset, Ratio=0.1).}

\label{imagenet}
\end{figure*}

We visualize our synthetic dataset in Figure \ref{syn_img}, the synthetic dataset is the result on CIFAR10 with IPC=10 by three methods. Figure \ref{syn_img}a shows the synthetic dataset distilled by Discrete Wavelet Transform, each line represents a category, because we randomly sampled some high-frequency information and low-frequency information each time for fusion in the process of distillation using DWT, so it makes the distillation process can be better learned the high-frequency information, and the visualization result proves this. Figure \ref{syn_img}b shows the results of the synthetic dataset distilled by the Learnable Discrete Cosine Transform (LDCT). Compared with DWT, the LDCT pays more attention to the high-frequency information, which further argues that the high-frequency information is very conducive to the learning of the synthetic dataset. Figure \ref{syn_img}c shows the visualization results of our spectral decomposition method, and it can be found that our method not only focuses on the low-frequency information well but also makes better use of the high-frequency information, which makes the synthetic dataset better acquire the information on different frequency bands. Figure \ref{svd} shows the visualization results of SVD and LSVD on CIFAR10, respectively, and Figure \ref{cifar100}, Figure \ref{tiny}, Figure \ref{imagenet} shows the visualization results of our method on CIFAR100, TinyImageNet, ImageNet Subset respectively, which clearly further demonstrates that our method can make better use of the spectral information and improve the compression rate at the same time. It is worth noting that our synthesized data is difficult to identify which category it belongs to, and thus can be well applied to privacy protection. 

